\documentclass[runningheads]{llncs}
\usepackage{graphicx}
\usepackage{comment}
\usepackage{amsmath,amssymb} 
\usepackage{color}
\usepackage[linesnumbered,ruled,vlined]{algorithm2e}
\SetKwInOut{KwInput}{Input}
\SetKwInOut{KwOutput}{Output}
\DeclareMathOperator*{\argmax}{arg\,max}
\DeclareMathOperator*{\argmin}{arg\,min}

\begin{document}
\pagestyle{headings}
\mainmatter
\def\ECCVSubNumber{1942}  

\title{Efficient Transfer Learning via Joint Adaptation of Network Architecture and Weight} 


\titlerunning{Transfer Learning via Joint Adaptation of Network Architecture and Weight}
%
\author{Ming Sun\inst{1}\orcidID{0000-0002-5948-2708} \and
Haoxuan Dou\inst{1}\orcidID{0000-0002-0237-6402} \and
Junjie Yan\inst{1}}
\authorrunning{M. Sun et al.}
%
\institute{SenseTime Group Limited, Beijing, China}
\maketitle

\begin{abstract}
Transfer learning can boost the performance on the target task by leveraging the knowledge of the source domain.
%
Recent works in neural architecture search (NAS), especially one-shot NAS, can aid transfer learning by establishing sufficient network search space.
However, existing NAS methods tend to approximate huge search spaces by explicitly building giant super-networks with multiple sub-paths, and discard super-network weights after a child structure is found.
Both the characteristics of existing approaches causes repetitive network training on source tasks in transfer learning. 
%
%
To remedy the above issues, we reduce the super-network size by randomly dropping connection between network blocks while embedding a larger search space.
Moreover, we reuse super-network weights to avoid redundant training by proposing a novel framework consisting of two modules, the neural architecture search module for architecture transfer and the neural weight search module for weight transfer.
These two modules conduct search on the target task based on a reduced super-networks, so we only need to train once on the source task.
We experiment our framework on both MS-COCO and CUB-200 for the object detection and fine-grained image classification tasks, and show promising improvements with only $O(C^{N})$ super-network complexity.
%


\end{abstract}

\begin{figure}[t]
\begin{center}
\includegraphics[scale=0.3]{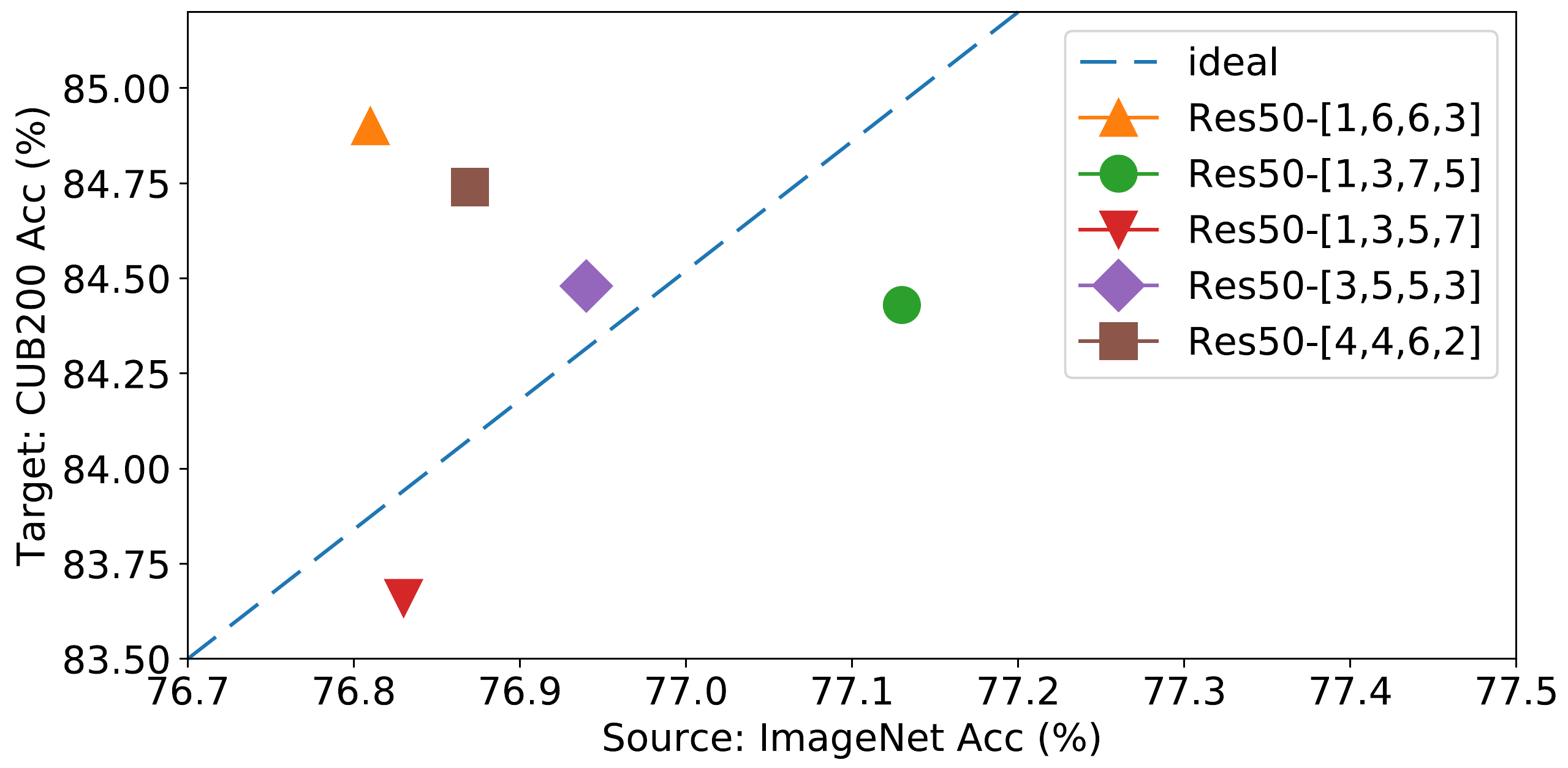}
\end{center}
  \caption{The plot shows the performances on target and source tasks for models of same size but different architectures. The numbers inside the brackets behind each \textit{ResNet-50} in the legend are the block numbers for each of the four stages of ResNet. The five ResNet-50 models have different block allocations but the same model size.} 
\label{fig:intro}
\end{figure}

\section{Introduction}

Deep neural networks have achieved significant successes in computer vision tasks like image classification \cite{se-block,resnext,bagoftricks,efficentnet}.
However, the success of deep networks tend to highly depend on a large amount of training data to ensure optimal training\cite{instagram}.
Therefore, insufficient training data can be an inescapable issue for tasks without a huge dataset, such as segmentation, object detection or medical image analysis.
Deep transfer learning was proposed to alleviate the data insufficiency by leveraging a massive source datasets to assist training on the target tasks\cite{survey}.
Most existing deep transfer learning methods apply transfer on networks of fixed architectures, with regularization on instance, feature or weight spaces\cite{survey,survey2}, and the result of transfer learning is reflected only through the model weights.
We argue that the architecture engineering also plays a vital role in the outcome of the knowledge transfer.
We perform direct finetuning on ImageNet\cite{imagenet} pre-trained ResNet-50\cite{resnet} models of different architectures on a fine-grained classification dataset CUB-200-2011\cite{WahCUB_200_2011}.
We plot their performances on source and target tasks in Fig.\ref{fig:intro}. We see that far from the ideal that better source models leads to better target models, when the model architecture is taken into consideration we see certain architecture is related to higher performance on target while suffering on source compared to models of different architecture, such as the yellow triangle in Fig.\ref{fig:intro}.
Therefore we argue that the potential of architecture engineering in the transfer learning process can be great, which is also recognized and explored in recent literature \cite{crnet}.
%
%
%

For architecture engineering, neural architecture search (NAS) methods have shown promising results.
NAS methods can be roughly divided into two categories: reinforcement learning (RL) based and gradient based \cite{oldrl1,oldrl3,mnas,darts,fbnet,proxylessnas}.
RL based NAS methods tend to be computationally expensive and not naturally fit for transfer learning, while gradient based methods, in particular single-shot NAS methods show robust and efficient search process \cite{singleshotnas,singleshotnas2,crnet}.
Single-shot NAS methods often use a super-network structure, which is a giant network subsuming a great collection of child network structures.
These methods provide fast architecture search and one may propose to utilize them for transfer learning.
However, we argue that existing methods cause inefficient transfer due to two reasons.
First,the super-networks employed in previous literature tend to be massive in size and result in slow training, which is insufficient since normally in transfer learning the size of the source task dataset is also huge.
Moreover, existing approaches discard super-networks after an ideal child architecture is found and child models are used to be retrained.
While in transfer learning discarding the super-network weights is not economical since the super-network weights can be inherited onto the child model for weight transfer.

To amend the aforementioned issues, we propose to first reduce the super-network size. 
Inspired by \cite{crnet} we use the allocation of networks among stages as our search space.
But unlike the multi-path super-network utilized in \cite{crnet}, we adopt a single-path super-network, which in-explicitly embeds a rich search space by sharing blocks among potential paths.
Specifically, each potential path can be obtained by dropping blocks in different stages of the super-network.
We train it by randomly dropping connection between network blocks during training iterations.
In this way we create a super-network of reasonable size yet with a rich search space, and therefore limit the computational expenses for training on source.
For instance, the super-network utilized in \cite{crnet} is equivalent to a ResNet-923 \cite{resnet}, while the reduced version is only equivalent to a ResNet-182 while embedding a richer search space.
Second, we aim to reuse the super-network weight by proposing a framework consisting of two modules: the neural architecture search module and the neural weight search module.
In the neural architecture search module, given the trained super-network we conduct architecture transfer by greedily searching for the target structure on target.
For the neural weight search module, the target structure reuse the network weights inherited from the super-network and fine-tune on target.
By reusing weights from the super-network, we avoid repetitive retraining.
Combining these two modules, we are able to effectively incorporate architecture engineering into the transfer learning process.

Our contributions are as following:
\begin{itemize}
    \item We demonstrate that the network architecture is crucial in the outcome of transfer learning, and therefore propose to incorporate architecture engineering into the pipeline of transfer learning for modern computer vision tasks such as image classification, object detection and instance segmentation.
    \item We propose a novel transfer learning framework, which adopts a single-path super-network for fast source training and incorporates both architecture and weight transfer for effective and fast transfer learning.
    \item Our experiments on various tasks including object detection and fine-grained classification show that our framework's robustness to diverse tasks. Moreover, our experiments on segmentation shows the good transfer ability of target models our method generates. Our method is able to boost the model performance on these tasks while keeping almost the same FLOPs.
\end{itemize}{}
%

%




\section{Related work}

\subsection{Transfer Learning}
Transfer learning addresses the problem of training with insufficient training data on a target task, by leveraging a massive dataset from a source domain\cite{survey2,survey}. Transfer learning focuses on what and how to transfer between the source and target domain, and different methods aim to address the two concerns in different forms such as the feature space, the instance space or the model weights. In terms of the model weights, one popular method called fine-tuning is to directly adapt the network pre-trained on a large scale source dataset to the target domain, or take the pre-trained network as backbone and add high-level layers for different target tasks such as recognition\cite{instagram,better_imagenet,fine_grain,sun2018multi}, object detection\cite{faster-rcnn,detection}, and segmentation\cite{maskrcnn}.
This method is shown to be more effective than a randomly initialized networks\cite{pretrain}.
On the hand, transfer learning methods also utilize regularization on the instance space and feature space to promote efficient transfer from source to target, by either re-weighting or re-sampling data from source domain to aid target domain learning \cite{boosting1,boosting2,boosting3,ensemble_resampling,metriclearningreweighting}, or by regularizing target domain learning through minimizing distance between the feature spaces of target and source\cite{mmd1,mmd2,mmd3,TCA}. Moreover, adversarial learning is adopted to create domain-invariant models for robust transfer learning\cite{adv1,adv2,adv3,adv4}. Recently, \cite{jang2019learning} propose to use meta-learning to do transfer learning between networks of heterogeneous structures and tasks, by learning a meta-model deciding what layer and feature should be paired for transfer.

However, the aforementioned methods mostly care about transferring knowledge between networks of fixed architectures. We instead incorporate the network architecture as a variable in the transfer process, to allow the target model to adapt its architecture to the target task.

\subsection{Neural Architecture Search}
Neural architecture search methods search network architecture on a fixed task, normally an image classification task\cite{nasnet}\cite{mnas}. Early NAS methods often conduct search in a nested manner, where numerous architecture is sampled from a large search space and trained from scratch, during which reinforcement learning\cite{oldrl1,oldrl3,mnas} or evolution\cite{evolution,evolution2,evolution3,evolution4} are used. These methods usually require a giant amount of computational resources. Recent NAS methods adopt the weight-sharing protocol to reduce computational intensity by leveraging a super-network which subsume all architecture\cite{weightshare,proxylessnas,fbnet,darts,snas}. In particular one-shot NAS methods train the super-network with stochastic path\cite{dropblock,singleshotnas,stochastic_path,dropblock}, and then search for the optimal architecture from the trained super-network in a separate step\cite{singleshotnas,singleshotnas2,crnet}. They normally use super-network with multiple paths, different paths consist of different kernel sizes, dilation ratios, channel numbers and block allocations in order to achieve good adaption to fixed tasks.  Even greatly reduced in recent works\cite{crnet}, the search spaces of recent NAS methods are still large and fine-grained. We instead use a single long path with skip connection to in-explicitly embed a large collection of possible sub-paths, which limits our complexity compared to existing methods. 

\begin{figure*}[t]
\begin{center}
\includegraphics[scale=0.35]{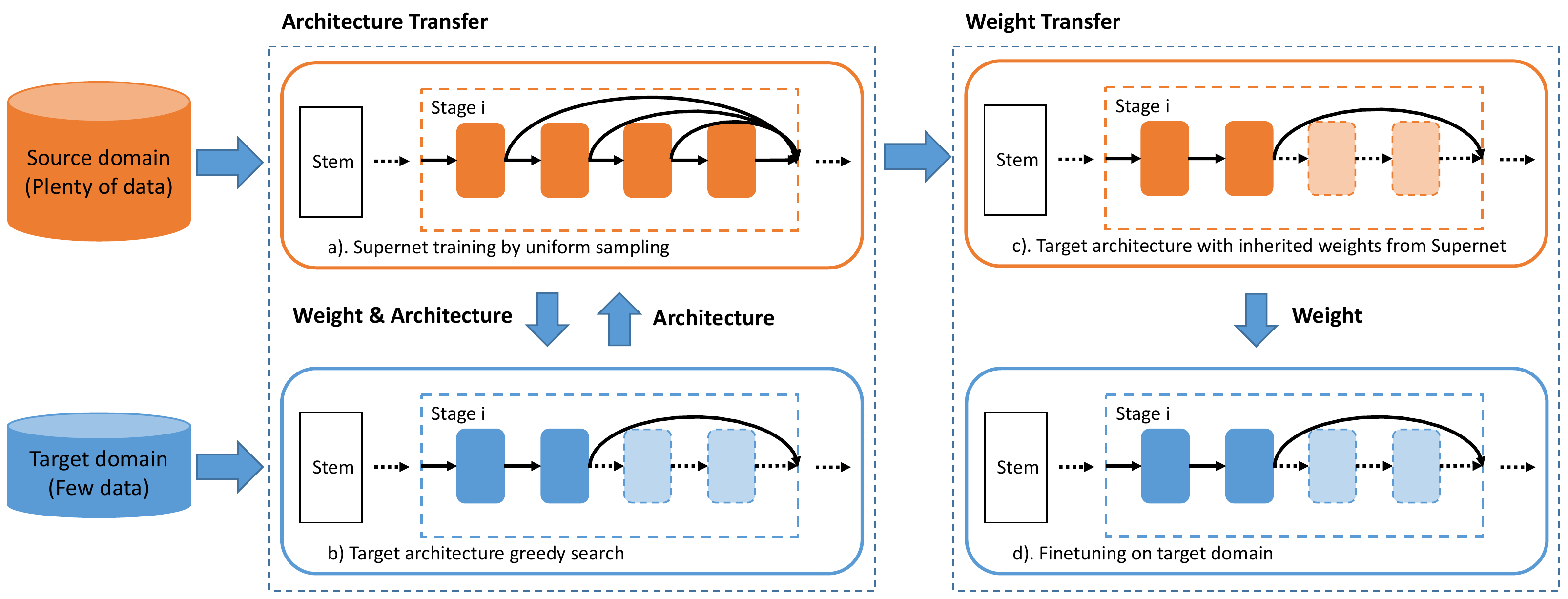}
\end{center}
  \caption{Our transfer learning framework consists of the architecture transfer and the weight transfer. For the architecture transfer on the left part of the figure, the super-network is trained on source domain and passed to target domain for the greedy search, and the searched architecture is passed back to inherit weight from the super-network. For the weight transfer network on the right, the target architecture with inherited weights is fine-tuned on source and target.}
\label{fig:method}
\end{figure*}

\section{Method}

In this section, we first introduce our problem setting, and then describe the super network, which provides a strong source model for later transfer. We then introduce the neural architecture search module to optimize the structure of the target model and the neural weight search module for weight transfer. Finally we demonstrate the potential of our transferred neural network to combine with various network blocks or hand-designed modules. The overview of our method is shown in Fig.\ref{fig:method}

\subsection{Problem Setting}
We consider transfer learning from the source domain and task \{$\mathcal{D}_s, \mathcal{T}_s$\} to the target domain and task \{$\mathcal{D}_t, \mathcal{T}_t$\}, where the target can come from a diverse set of domains.
We follow the definitions from \cite{survey,survey2} and denote domain as $\mathcal{D} = \{\mathcal{X}, P(X)\}$ with $\mathcal{X}$ as the data space and $P(X)$ as the conditional probability where $X = \{x_i, \dots, x_n\} \in \mathcal{X}$ is the domain data. We also define task as $\mathcal{T} = \{\mathcal{Y},f(\cdot)\}$ with $\mathcal{Y}$ as the label space and $f(\cdot)$ being an objective predictive function that maps $x \in X$ to $y \in \mathcal{Y}$ and is learnt during training.
In transfer learning in general, $\mathcal{D}_s \neq \mathcal{D}_t$, $\mathcal{T}_s \neq \mathcal{T}_t$, and the size of source data is usually much larger than the size of the target data, that is $|D_s| \gg |D_t|$.

We define the backbone network by $\mathcal{N}(\phi, \mathit{w_\phi})$ as a transformation from the data space to the feature space, where $\phi \in \mathcal{A}$ denotes the model architecture and $\mathit{w}_\phi \in \mathcal{W}_\mathcal{A}$ denotes the model weight. $\mathcal{A}$ and $\mathcal{W}_\mathcal{A}$ respectively define the architecture search space and corresponding weight space.
We aim to find a model $\mathcal{N}(\phi_t, \mathit{w}_{\phi_t})$ on target to maximize performance on target domain validation set given the source \{$\mathcal{D}_s, \mathcal{T}_s$\},
\begin{equation}
    (\phi_t, \mathit{w}_{\phi_t}) = \argmin_{(\phi,\mathit{w}_{\phi}) \in (\mathcal{A}, \mathcal{W}_\mathcal{A})} \mathcal{L}^{target}_{val}(\mathcal{N}(\phi, \mathit{w}_{\phi}))
\end{equation}
where $\mathcal{L}_{val}$ represents the validation loss on the target domain.

Traditional transfer learning methods focus on optimizing the model weight $\mathit{w}_{\phi_t}$ with a fixed $\phi_t$ by training with constraints on feature space, instance space or by inheriting model weights and fine-tuning, that is
\begin{equation}
    \min_{\mathit{w}_{\phi_t} \in  \mathcal{W}_\mathcal{A}} \mathcal{L}^{target}_{val}(\mathcal{N}(\phi_t, \mathit{w}_{\phi_t}))
\end{equation}
Here unlike traditional transfer learning methods we incorporate neural architecture search and can turn Eq.1 into a bi-level optimization problem,
\begin{equation}
\begin{aligned}
     \min_{\mathit{w}_{\phi_t} \in  \mathcal{W}_\mathcal{A}} \mathcal{L}^{target}_{val}(\mathcal{N}(\phi_t, \mathit{w}_{\phi_t}))\\
    \text{s.t. } \phi_t = \argmin_{\phi \in \mathcal{A}} \mathcal{L}^{target}_{val}(\mathcal{N}(\phi, \mathit{w}^s_{\phi}))
\end{aligned}
\end{equation}
where $\mathit{w}^s_{\phi}$ is the optimized weights on source given a super-network architecture $\phi$.
As we introduce NAS into the process of transfer learning, it becomes non-trivial to accommodate NAS approaches for efficient transfer.
In particular we need our architecture search step to take the difference between sizes of source and target datasets into consideration during transfer learning.

\subsection{Source Super-Network Training}
The ideal super-network for transfer learning should contain three qualities.
Firstly, the super-network should embed a rich search space for finding a powerful network structure.
Secondly, the super-network size should be small for efficient training.
Finally, the super-network should have the similar network hierarchy as the potential child model for weight sharing.
Previous NAS approaches adopting weight sharing protocols \cite{proxylessnas,fbnet,darts,singleshotnas,singleshotnas2} often establish giant super-networks with complex structures, which violates the second quality.
In addition, the search space they embed tend to be complicated.

Inspired by recent work on single-shot NAS \cite{crnet}, we utilize the allocation of network blocks among different stages as the search space for its efficacy and simplicity.
However, we find that the multi-branches of different numbers of blocks often utilized in these one-shot NAS methods cause the model size to significantly increase with depth.
We instead just use one branch of the maximum amount of blocks, which contains all possible paths as its sub-paths.
By sharing blocks among potential paths, we reduce the super-network size while maintaining a powerful search space.
%

To sufficiently train the super-network such that each child model represented by its sub-paths can provide robust weight for neural architecture search on target, we randomly drops networks blocks from the forward and backward pass during training.
Specifically, a stage in the super-network consists of $N$ blocks in a sequential order.
During each iteration in training we uniformly sample $S$ from $[1,N]$, and only keep the first $S$ block for this round of training.
In this way, different combination of sub-paths are trained and excessive co-adaptation between blocks are avoided.
The super-network is able to still embed a powerful search space while being small in size.

\subsection{Neural Architecture Search on Target}

Given the super-network from the source and the required model size constraint $C$, we first fine-tune the super-network on the target task and then search for a robust architecture.
We first define our super-network and the corresponding search space.
We can represent the super-network architecture as $\phi_s = [N^s_1, \cdots, N^s_{ns}]$ where $N_i$ refers to the number of blocks at the $i_{th}$ stage. Normally the number of stages $ns = 4$ for most of networks.
For instance, we can define ResNet50 as $\phi_{res50} = [3, 4, 6, 3]$.
With the super-network and the required model size constraint $C$, we can define the architecture search space under model size $C$ as,
\begin{equation}
        \mathcal{A}_C = \{\phi_t| \phi_t \in \Gamma(\phi_s), sum(\phi_t) = C\}
\end{equation}
where $\Gamma(\phi_s) = \{\phi_t = [N^t_1,\cdots,N^t_{ns}] | N_i^t \in [1,\cdots,N_i^s]\}$ denotes the search space of all possible child models.


%
Random search or evolution are often applied for specified computation limitation.
However, instead of having a fixed computation limits like previous approaches often do, we aim to search for optimal allocation for all possible model sizes.
Here we make the assumption that given a model size $c$ and the corresponding optimal target architecture $\phi^*_c$, the optimal architecture $\phi_{c+1}$, corresponding to model size $c+1$, contains $\phi_c$ as a sub-graph.
Here we denote the search space of $\phi_{c+1}$ conditional on an optimal $\phi^*_c$ as $\mathcal{A}^*_{c+1}$,
\begin{equation}
    \mathcal{A}^*_{c+1} = \{\phi_{c+1}|\phi^*_{c} \subset \phi_{c+1}, \phi_{c+1} \in \mathcal{A}_{c+1} \}
\end{equation}

Based on this greedy assumption, we can inductively search for optimal target network architecture $\phi_t$ starting from the minimal architecture $\phi_{ns} = [1,\cdots,1]$, where $ns$ is the number of stages. 
In the inductive search step, given optimal $\phi_c$, we find $\phi_{c+1}$ by adding one more block that maximize the resulting model's performance on the target validation set, that is
\begin{equation}
    \phi^*_{c+1} = \argmax_{\phi_{c+1} \in \mathcal{A}^*_{c+1}} \mathcal{L}_{val}(\mathcal{N}(\phi_{c+1}, \mathit{w}^s_{\phi_{c+1}}))
\end{equation}
$\phi^*_{c+1}$ can be found by running network evaluation for $ns$ times, where the evaluated network is obtained by appending the next block to its path at each stage.
We repeat the induction step for several times until we reach the model size constraint $C$, and we have the optimal architectures for models whose sizes ranges from $ns$ to $C$.
%
%
%
%
In detail, this algorithm is shown in Alg.\ref{algo:search} and left bottom of Fig.\ref{fig:method}.

\subsection{Neural Weight Search on Target}
After we have a optimized neural architecture $\phi_t$, we focus on the transfer of network weights.
Existing NAS methods \cite{crnet,singleshotnas,singleshotnas2} often discard the network weights after search and retrain, which is time-consuming.
We reuse the weight inherited from the super-network through the neural weight search, which save time used in repetitive training. 
Specifically for the neural weight search we apply the fine-tuning method often employed in the transfer learning.
Given the searched target network \{$\phi_t$, $w^s_{\phi_t}$\}, where $w^s_{\phi_t}$ is directly inherited from the source super-network, we first fine-tune it on the source domain to obtain robust source network weights, and then we fine-tune the resulting network on the target domain to get \{$\phi_t$, $w_{\phi_t}$\}. 

In this way, we ensure a sufficient transfer of network weight based on the target architecture.
The process is shown in the right part of Fig.\ref{fig:method}.
Note that the weight transfer step can be extended to use other knowledge transfer methods beyond weight fine-tuning, such as knowledge distillation \cite{knowledge} or feature mimicking\cite{mimicking}. 

\begin{algorithm}[t]
\label{algo:search}
\SetAlgoLined
\KwInput{Model size constraint $C \geq ns$, super-network $\mathcal{N}(\phi_s, \mathit{w}_s)$}

\KwOutput{Target architecture $\phi_t^*$ with $sum(\phi_t) = C$}
 initialize start network $\phi_{ns} = [1,\cdots,1]$, $c = ns+1$\;
 \While{$c \leq C$}{
 $\phi^*_c = \argmax_{\phi_{c} \in \mathcal{A}^*_{c}} \mathcal{L}_{val}(\mathcal{N}(\phi_{c}, \mathit{w}_{\phi_{c}}))$\;
 $c = c + 1$\;
 }
\Return{$\phi_t^* = \phi_C^*$}
\caption{Greedy Block Search}
\end{algorithm}

\subsection{Generalization over Diverse Structures}

\subsubsection{Generalization over diverse blocks}

%
%
Our transfer learning framework can generalize to a diverse collection of network blocks, hand-designed or generated using NAS.
We first experiment on blocks of the widely used hand-designed MobileNetV2\cite{mobilenetv2} and ResNet\cite{resnet}. 
Moreover, we apply our method using blocks of MnasNet-b0\cite{mnas}, which is generated through neural architecture search using reinforcement learning on ImageNet\cite{imagenet}.
Table.\ref{table:od} shows that on detection task, our framework transfers well on all three network blocks for the object detection task with a universal performance improvement.

%
%
%


\subsubsection{Generalization over hand-designed modules}
Our framework also generalizes well over network blocks enhanced by hand-designed modules. We apply our method to two widely recognized modules.
The Squeeze-Excitation module\cite{se-block} improves network performances through a channel-wise attention mechanism, while ResNeXt\cite{resnext} achieves improvements by multi-branching a ResNet block in order to increase the computational cardinallity.
We show the robustness of our method to block enhancements by enhancing a transferred ResNet-50 model with the two modules described above on object detection task.
As is shown in Table.\ref{table:od}, the transferred ResNet-50 architecture still has 1 point improvement in mAP on the basis of SE-block and ResNeXt.

\section{Experiment}
For all our experiments we implement our methods using the Pytorch framework\cite{pytorch}, and in this section we describe the tasks we experimented on and the corresponding results.

\begin{table}[t]
\caption{Performance of different network configures for object detection on MS-COCO\cite{lin2014microsoft} dataset using Faster-RCNN with FPN\cite{ren2015faster,FPN}.}
\begin{center}
\begin{tabular}{|l|c|c|}
\hline
Method & mAP &  FLOPS \\
\hline\hline
ResNet50 \cite{crnet} & 36.4 & 3.991G \\
CR-ResNet50 \cite{crnet} & 37.4 & 3.991G \\ 
Transfer-ResNet50 & 37.8 & 3.991G \\ 
ResNet101 & 38.5 & 7.62G \\
CR-ResNet101 \cite{crnet} & 39.5 & 7.62G \\ 
Transfer-ResNet101 & 39.8 & 7.62G \\ 
MobileNetV2 & 32.2 & 312.34M \\ 
CR-MobileNetV2 \cite{crnet} & 33.5 & 329.21M \\ 
Transfer-MobileNetV2 & 34.0 & 311.68M \\ 
\hline
MnasNet-b0 & 34.2 & 313.14M \\
Transfer-MnasNet-b0 & 34.7 & 318.37M \\
\hline
SE-ResNeXt50 & 38.9 & 4.130G \\
Transfer-SE-ResNeXt50 & 39.9 & 4.132G \\
Grid-RCNN-ResNet50 & 39.5 & 3.991G \\
Gird-RCNN-Transfer-ResNet50 & 40.3 & 3.991G \\
Cascade-MaskRCNN-ResNet101 & 43.3 & 7.62G \\
Cascade-MaskRCNN-Transfer-ResNet101 & 44.1 & 7.62G \\

\hline
\end{tabular}
\end{center}
\label{table:od}
\end{table}

\subsection{Objection Detection}
For the object detection task, we conduct experiments on the MS-COCO dataset \cite{lin2014microsoft}, which contains 118K training 
images and 5K validate images (\textit{minival}) for 80 classes.
%
%
The evaluation metric is the mean average precision (mAP). 

We conduct experiments on three types of network blocks: the ResNet\cite{resnet} bottleneck block, the MobileNetV2\cite{mobilenetv2} block and the MnasNet\cite{mnas} block.
For the super-network consisting of ResNet bottleneck blocks\cite{resnet}, we set the block allocation to be $T_s = [8, 10, 36, 14]$, whose FLOPs is roughly equal to that of ResNet-200.
For the super-network consisting of MobileNetV2 block\cite{mobilenetv2}, we set the block allocation to be $T_s = [5, 6, 8, 7, 7]$ with 5 stages.
For the super-network consisting of MnasNet blocks\cite{mnas}, the configuration is same to that of MobileNetV2.

We use the ImageNet dataset\cite{imagenet} as the source domain and train our super-networks on it for 100 epochs with label smoothing of $0.1$ and 5 epochs of warmup.
We use 32 GPU cards with a total batch size of 2048, and the base learning rate is $0.01$.
For the learning rate we conduct the step decay scheme, where we drop the learning rate by $10$ at epoch of 30, 60, 80.
The weight decay is set to be 0.0001.

For the object detection task on MS-COCO dataset, we adopt Faster-RCNN\cite{faster-rcnn} in combination with the feature pyramid network (FPN)\cite{FPN}.
For finetuning, the base learning rate and weight decay are set to be 0.04 and 0.0001 respectively.
We use 32 GPU cards with a total batch size of 64.
Note that different to ResNet models, where the network blocks at the adjacent stages share the same FLOPs, the MobileNetV2 models' blocks need to be re-weighted during the architecture search.
Specifically, the block number at each stage is re-weighted according to that stage's computation FLOPs, and then we do the greedy search until the sum of the re-weighted block numbers reaches the computational limits.
We apply the same re-weighting to the architecture search with MnasNet, too.

%

{\bf Results on Different Network Blocks.}
To demonstrate the effectiveness of our proposed method, we conduct transfer learning on ResNet, MobileNetV2 and MnasNet-b0 with configurations described above.
The results are shown in Table.\ref{table:od}.
For ResNet, we search the best network architecture for the popular ResNet-50, which has 16 blocks in total.
The best architecture is $[1, 3, 7, 5]$, which get 37.8\% mAP and 1.4\% higher than default $[3, 4, 6, 3]$ configuration on MS-COCO.
We also transfer the ResNet-101 model, and our transferred model reach 39.8\% mAP surpassing the default configure by 1.3\%, with the optimal architecture as $[2,5,19,7]$.
For the MobileNetV2 and the MnasNet-b0, their searched best architectures are $[2, 2, 2, 3, 5]$ and $[3, 3, 2, 3, 4]$ respectively.
Also, their performances are better than baselines' on MS-COCO \textit{minival} while keeping almost the same backbone FLOPs.
Furthermore, we show better results compared to \cite{crnet} on all three network blocks above.
We think the reason for the better results lie in the richness of our search space, as our search space contains all sub-structures fulfilling the model size constraint. 
However, the search space adopted in \cite{crnet} skips certain sub-structures since paths with certain number of blocks are not built in the super-network to prevent its size from being computationally unbearable.

Moreover, we show that the boosted SE-ResNeXt-50, enhancing the transferred ResNet-50 blocks with multi-branching\cite{resnext} and SE-module\cite{se-block}, also shows solid performance improvement on MS-COCO \textit{minival}, compared to the original SE-ResNeXt-50.
Furthermore, we show that our transferred ResNet backbones also outperform their default counterparts when combined with the Grid-RCNN framework\cite{gridrcnn} and the Cascade-MaskRCNN framework\cite{cascadercnn}, with both 0.8\% improvements in mAP on the MS-COCO \textit{minival}.

These results demonstrate the robustness of our methods over various types of network blocks with or without hand-designed module or strong frameworks for enhancements.

\begin{table}[t]
\caption{Transferred ResNet on MS-COCO. The colored numbers indicate that adding one block respectively in the third(red \& blue) and fourth(green) stages seems to increase performance the most on small(red), medium(blue) and large(green) objects.}
\begin{center}
\begin{tabular}{|l|c|c|c|c|c|c|}
\hline
Method & mAP & Block & FLOPS(G) & Box$_s$ & Box$_m$ & Box$_l$ \\
\hline\hline
Transfer-ResNet37& 36.4 & [1, 3, \textcolor[rgb]{1,0,0}{4}, 4]& 3.138 & \textcolor[rgb]{1,0,0}{0.2089} &0.3987&0.4747  \\
Transfer-ResNet41 & 36.7 & [1, 3, \textcolor[rgb]{1,0,0}{5}, \textcolor[rgb]{0,1,0}{4}] & 3.352& \textcolor[rgb]{1,0,0}{0.2178} &0.4031&\textcolor[rgb]{0,1,0}{0.4608} \\
Transfer-ResNet44 & 37.0 & [1, 3, \textcolor[rgb]{0,0,1}{5}, \textcolor[rgb]{0,1,0}{5}] & 3.565 & 0.2128 &\textcolor[rgb]{0,0,1}{0.4051}&\textcolor[rgb]{0,1,0}{0.4805} \\
Transfer-ResNet47 & 37.4 & [1, 3, \textcolor[rgb]{0,0,1}{6}, 5] & 3.777 & 0.2153 &\textcolor[rgb]{0,0,1}{0.4117}&0.4836 \\
Transfer-ResNet50 & 37.8 & [1, 3, 7, 5] &  3.991& 0.2158 &0.4180&0.4925 \\
Transfer-ResNet53 & 38.0 & [1, 3, 7, 6] & 4.205& 0.2168 &0.4148&0.4958 \\
\hline
\end{tabular}
\end{center}
\label{table:differentsize}
\end{table}

{\bf Results on Different Computation Constraints.}
Apart from the common network architecture configures like ResNet-50, we also explore the potential of our framework on a series of target model architectures optimized under different computational constraints, such as ResNet-37, ResNet-41 and so on.
With the pre-trained super-network, we can search for models of different size on the target task by assigning different computation requirements.
For example, you can set 15 for searching a best ResNet47 network configuration.
Also, thanks to our greedy search algorithm, we can efficiently search for models of larger size on the basis of previously transferred smaller-sized model without starting from scratch.
%
%
On Table.\ref{table:differentsize}, we show some the ResNet configurations of different computational complexity and their corresponding performances on MS-COCO \textit{minival}. 
For example, $T^*_{res47} = [1,3,6,5]$, $T^*_{res50} = [1,3,7,5]$, $T^*_{res53} = [1,3,7,6]$ are generated from the super-network. 
%

\begin{figure}[t]
\begin{center}
\includegraphics[scale=0.34]{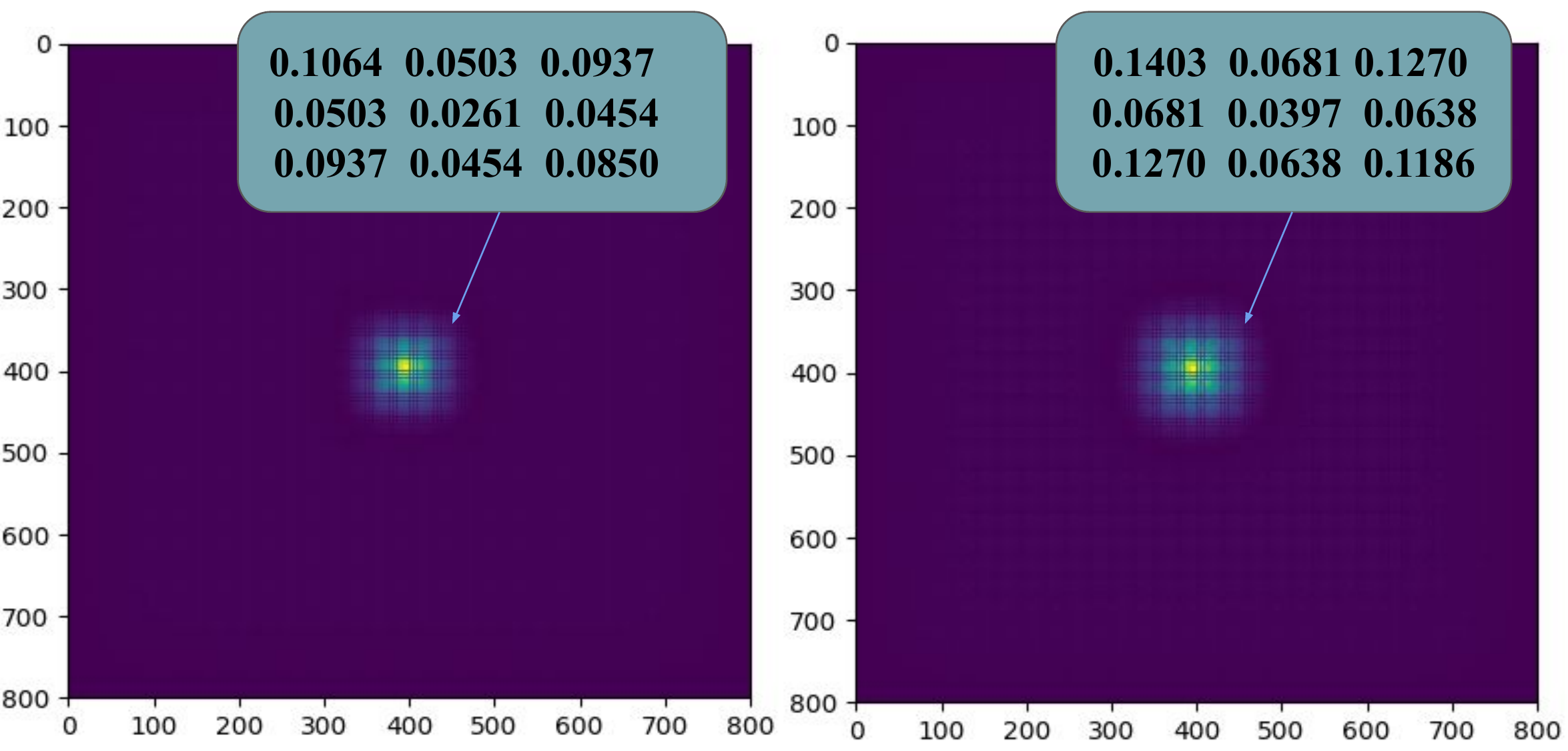}
\end{center}
  \caption{The effective receptive fields of the baseline network and our transferred network from the last convolution layer. The left part is the baseline and the right part is our transferred network.}
\label{fig:erf}
\end{figure}

{\bf Visualization.}
To further understand the transferred backbone, 
we visualize the backbone effective receptive filed (ERF), which is the key to the objection detection task.
Based on the method proposed  in \cite{luo2016understanding}, the receptive fields of the center neuron on the last convolution layer are visualized in Figure.\ref{fig:erf}. 
In detail, the figure is generated by setting input values to 1 and propagating using the neuron in the center
of the last convolution layer. 
ReLU operations are abandoned to better visualize the intensity of connections.
As shown in the Figure.\ref{fig:erf}, 
the ERF size of our transferred network is larger than that of the baseline network.
We also calculate the outer response number, as shown in the green box on right top of the ERF figures.
Except the stronger intensities of center region, the intensities of outer region from the ERF of our transferred is also stronger than that of the default baseline network. 
The strong intensity of ERF is important for the object detection task which contains higher scale variance compared with the classification task, and our transferred network manages to capture this characteristics through the architecture engineering on the target task.

{\bf Complexity Analysis.} 
The one-shot NAS methods often employ super-networks made up of multi-branches while our super-network only keeps the longest path.
For our super-network architecture $[8, 10, 36, 14]$ with the block sum with $68$ is equal to the ResNet-206, while the equivalent one-shot super-network using multi-branches needs to contain 862 blocks to capture all the potential sub-paths. Since a ResNet bottleneck block contains three layers and the network stem contains two layers, the one-shot super-network is equivalent to a ResNet-2588.
Moreover the search space of regular one-shot methods needs to contain all combinations of block numbers of each stage such that the model size requirement is met, leading to a space of order $O(C^{ns})$. While our search space, thanks to the greedy search algorithm, is only of order $O(C\times ns)$


\begin{table}[t]
\caption{CUB-200-2011 fine-grained classification with 5799 training images.}
\begin{center}
\begin{tabular}{|l|c|c|c|}
\hline
Method & mAP & Block number & ImageNet \\
\hline\hline
ResNet50 \cite{yue2018compact} & 84.05 & [3, 4, 6, 3] & 76.72 \\
ResNet50+NL \cite{non-local} & 84.79 & [3, 4, 6, 3] & - \\
ResNet50+CGNL \cite{yue2018compact} & 85.14 & [3, 4, 6, 3] & - \\
Transfer-ResNet50 & \textbf{84.98}  & [1, 6, 6, 3] & 76.81 \\
Transfer-ResNet50+NL \cite{non-local} & \textbf{85.42}  & [1, 6, 6, 3] & - \\
\hline
Transfer-ResNet50 & 84.95  & [2, 4, 7, 3] & \textbf{77.06} \\
Transfer-ResNet50 & 84.76 & [2, 3, 7, 4] & 76.78 \\
Transfer-ResNet50 &  84.74 & [4, 4, 6, 2] & 76.87 \\
Transfer-ResNet50 & 84.64 & [1, 4, 8, 3] & 76.93 \\
\hline
\end{tabular}
\end{center}
\label{table:cub2011}
\end{table}

\begin{table}[t]
\caption{CUB-200-2010 fine-grained classification with total 3000 training images.}
\begin{center}
\begin{tabular}{|l|c|c|c|}
\hline
Method & mAP & Block number & ImageNet \\
\hline\hline
ResNet50 & 68.01 & [3, 4, 6, 3] & 76.72 \\
Transfer-ResNet50 & \textbf{71.35}  & [1, 6, 6, 3] & 76.81 \\
\hline
Transfer-ResNet50 &  68.02 & [3, 5, 5, 3] & 76.94 \\
Transfer-ResNet50 & 67.72  & [2, 4, 7, 3] & 77.06 \\
Transfer-ResNet50 &  67.66 & [1, 3, 7, 5] & \textbf{77.13} \\
Transfer-ResNet50 & 67.23 & [1, 4, 8, 3] & 76.93 \\
\hline
\end{tabular}
\end{center}
\label{table:cub2010}
\end{table}

\begin{table*}[t]
\caption{Instance Segmentation Results on MS-COCO. Here we show comparison between our transferred network and baseline networks. We also demonstrate higher performance over more complex networks such as Cascade-MaskRCNN in the object detection task as shown in Tab.\ref{table:od} }
\begin{center}
\resizebox{\textwidth}{15mm}{
\begin{tabular}{|l|c|c|c|c|c|c|c|c|c|}
\hline
Backbone & Seg & Seg$_s$ & Seg$_m$ & Seg$_l$ &  Box &  Box$_s$ & Box$_m$ & Box$_l$ \\
\hline\hline
ResNet50          & 33.9 & 17.4 & 37.3 & 46.6 &  37.6 &  21.8 &  41.2 & 48.9 \\
Ours & 34.8(+0.9) & 17.9 (+0.5) & 38.3(+1.0)  & 48.1(+1.5) & 38.6(+1.0) & 22.5(+0.7) & 42.2(+1.0) & 50.6(+1.7) \\
\hline
ResNet101 & 35.6 & 18.6 & 39.2 &  49.5 & 39.7 & 23.4 & 43.9 & 51.7 \\
Ours & 36.4(+0.8) & 18.7(+0.1) & 40.0(+0.8) &  50.6(+1.1) & 40.6(+0.9) & 23.5(+0.1) & 44.5(+0.6) & 53.4(+1.7) \\
\hline
MobileNetV2 &  30.6 & 15.3 & 33.2 & 44.1 &  33.1 &  18.8 &  35.8 &  43.3 \\
Ours & 31.8(+1.2) & 16.0(+0.7) & 34.6(+1.4)  & 43.8(-0.3) & 34.8(+1.7) & 19.7(+0.9) & 37.6(+2.8) & 45.6(+2.3) \\
\hline
\end{tabular}}
\end{center}
\label{table:seg}
\end{table*}

\subsection{Fine-grained classification.}
In the fine-grained image classification task, since the data collecting and labeling is time-consuming, fine-grained image classification task dataset is often small.
For example, the commonly used dataset is CUB-200-2011 \cite{WahCUB_200_2011}, which contains 200 classes and 5994 training images in total.


We experiment on CUB-200-2011 following \cite{yue2018compact} using the input size of $448$ with ImageNet\cite{imagenet} dataset as the source domain.
When finetuning on target, all models are trained for 100 epochs with an initial learning rate of 0.01 and a step decay scheme dropping the learning rate by 10 at the epoch of 30, 60, 80.
We use one GPU card with the batch size of 64 samples.

We show the performances of transferred models on both CUB-200-2011 and ImageNet datasets in Table.\ref{table:cub2011}, and we can see the best transferred network architecture is $[1, 6, 6, 3]$ with more than $0.9$ point higher than the baseline.
%
We also include the several other top performing transferred networks in Table.\ref{table:cub2011}.
Note that the top performing transferred network on the CUB-200-2011 task shows worse performances on the ImageNet dataset, compared to the other transferred model with the architecture $[2, 4, 7, 3]$ that achieve $77.06$ accuracy on ImageNet while performing less optimal on target.
Moreover, the rest of the transferred models also demonstrate the inconsistency between source and target performances when the models are of different architectures.
This fact shows that finding a suitable architecture for a target model can play a role in transfer learning as vital as the weight transfer, if not more.
In addition, we conduct transfer experiments using ResNet-50 with the non-local module, showing better results than both the original and compact generalized non-local baselines \cite{non-local,yue2018compact}.

We also conduct experiments on CUB-200-2010 \cite{WelinderEtal2010}, which is an older and smaller version of CUB-200-2011.
We report the performances both on CUB-200-2010 and ImageNet in Table.\ref{table:cub2010}.
We observe that the optimal transferred network architecture on CUB-200-2010 vary from that of CUB-200-2011, but we also observe a $60$\% overlap of the top $5$ performing architectures on CUB-200-2011 and CUB-200-2010.


\begin{figure*}[t]
\begin{center}
\includegraphics[scale=0.35]{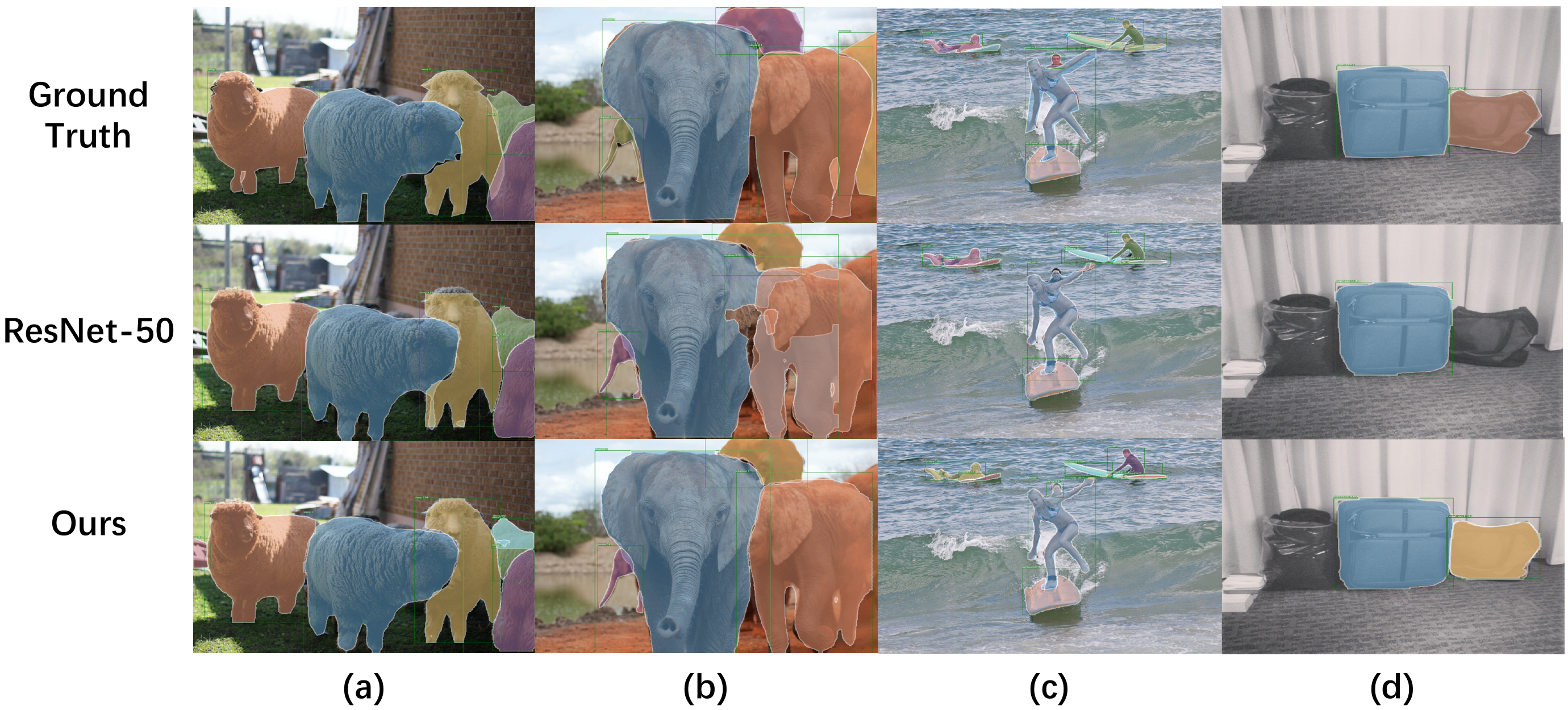}
\end{center}
  \caption{Some hard segmentation cases not handled well by baseline ResNet-50 but fixed by our transferred ResNet-50. a). Our model shows tighter instance masks. b). Our model shows more consistent masks. c). Our method successfully detects the small person behind the surfer, missed by the regular ResNet-50. d). Our models captures the bag on the right while the baseline misses it.}
\label{fig:seg}
\end{figure*}

\subsection{Semantic Segmentation.}
We wish to explore the potential of our transferred model to transfer further to other target tasks without modification to the architecture.
%
%
We conduct experiments on the MS-COCO instance segmentation task with the transferred models obtained from the object detection task.
In particular, we take the Mask-RCNN\cite{he2017mask} and replace the backbone with the transferred model, and then we directly finetune on MS-COCO instance segmentation dataset.

%
We report performances of our models in Table.\ref{table:seg}.
%
Our transfer-ResNet-50 with architecture of $[1, 3, 7, 5]$ outperforms its counterparts on both segmentation and detection with various object scales.
In Fig.\ref{fig:seg} we show some of the cases from MS-COCO dataset that the traditional ResNet-50 handles badly while our transferred model handles well.
%
For Transfer-ResNet-101, its segmentation and detection mAP are better than its default counterpart by  0.8\% and 0.9\%, and it also shows improvement on objects of all scales.
%
%
For Transfer-MobileNetV2, the mAP of detection and segmentation are better by 1.2\% and 1.7\% respectively.
Interestingly, the segmentation of larger object is inferior to baseline by 0.3\% while detection on larger object better by 2.3\%.
%
%

\subsection{Conclusion.}
In this paper, we propose a novel transfer framework containing a neural architecture search module and a neural weight search module.
In the architecture transfer we design a block-level search space and accordingly build an powerful super network on source, and search for the optimal architecture through a greedy algorithm on the target task. For the neural weight search module, we adopt weight fine-tuning, which can be smoothly replaced by other existing transfer learning methods to push the performance even higher.
Extensive experiments of our framework on various tasks show promising results .

\clearpage
%
%
\bibliographystyle{splncs04}
\bibliography{egbib}

\begin{thebibliography}{10}
\providecommand{\url}[1]{\texttt{#1}}
\providecommand{\urlprefix}{URL }
\providecommand{\doi}[1]{https://doi.org/#1}

\bibitem{oldrl3}
Baker, B., Gupta, O., Naik, N., Raskar, R.: Designing neural network
  architectures using reinforcement learning. CoRR  \textbf{abs/1611.02167}
  (2016), \url{http://arxiv.org/abs/1611.02167}

\bibitem{singleshotnas}
Bender, G., Kindermans, P.J., Zoph, B., Vasudevan, V., Le, Q.: Understanding
  and simplifying one-shot architecture search. In: Dy, J., Krause, A. (eds.)
  Proceedings of the 35th International Conference on Machine Learning.
  Proceedings of Machine Learning Research, vol.~80, pp. 550--559. PMLR,
  Stockholmsmässan, Stockholm Sweden (10--15 Jul 2018),
  \url{http://proceedings.mlr.press/v80/bender18a.html}

\bibitem{proxylessnas}
Cai, H., Zhu, L., Han, S.: Proxyless{NAS}: Direct neural architecture search on
  target task and hardware. In: International Conference on Learning
  Representations (2019), \url{https://openreview.net/forum?id=HylVB3AqYm}

\bibitem{cascadercnn}
Cai, Z., Vasconcelos, N.: Cascade {R-CNN:} delving into high quality object
  detection. CoRR  \textbf{abs/1712.00726} (2017),
  \url{http://arxiv.org/abs/1712.00726}

\bibitem{fine_grain}
Cui, Y., Song, Y., Sun, C., Howard, A., Belongie, S.: Large scale fine-grained
  categorization and domain-specific transfer learning. In: The IEEE Conference
  on Computer Vision and Pattern Recognition (CVPR) (June 2018)

\bibitem{boosting1}
Dai, W., Yang, Q., Xue, G.R., Yu, Y.: Boosting for transfer learning. In:
  Proceedings of the 24th International Conference on Machine Learning. pp.
  193--200. ICML '07, ACM, New York, NY, USA (2007).
  \doi{10.1145/1273496.1273521},
  \url{http://doi.acm.org/10.1145/1273496.1273521}

\bibitem{imagenet}
Deng, J., Dong, W., Socher, R., Li, L.J., Li, K., Fei-Fei, L.: {ImageNet: A
  Large-Scale Hierarchical Image Database}. In: CVPR09 (2009)

\bibitem{pretrain}
Erhan, D., Bengio, Y., Courville, A., Manzagol, P.A., Vincent, P., Bengio, S.:
  Why does unsupervised pre-training help deep learning? J. Mach. Learn. Res.
  \textbf{11},  625--660 (Mar 2010),
  \url{http://dl.acm.org/citation.cfm?id=1756006.1756025}

\bibitem{adv2}
Ganin, Y., Lempitsky, V.: Unsupervised domain adaptation by backpropagation.
  In: Proceedings of the 32Nd International Conference on International
  Conference on Machine Learning - Volume 37. pp. 1180--1189. ICML'15, JMLR.org
  (2015), \url{http://dl.acm.org/citation.cfm?id=3045118.3045244}

\bibitem{adv1}
Ganin, Y., Ustinova, E., Ajakan, H., Germain, P., Larochelle, H., Laviolette,
  F., Marchand, M., Lempitsky, V.: Domain-adversarial training of neural
  networks. J. Mach. Learn. Res.  \textbf{17}(1),  2096--2030 (Jan 2016),
  \url{http://dl.acm.org/citation.cfm?id=2946645.2946704}

\bibitem{dropblock}
Ghiasi, G., Lin, T.Y., Le, Q.V.: Dropblock: A regularization method for
  convolutional networks. In: Proceedings of the 32Nd International Conference
  on Neural Information Processing Systems. pp. 10750--10760. NIPS'18, Curran
  Associates Inc., USA (2018),
  \url{http://dl.acm.org/citation.cfm?id=3327546.3327732}

\bibitem{singleshotnas2}
Guo, Z., Zhang, X., Mu, H., Heng, W., Liu, Z., Wei, Y., Sun, J.: Single path
  one-shot neural architecture search with uniform sampling. CoRR
  \textbf{abs/1904.00420} (2019), \url{http://arxiv.org/abs/1904.00420}

\bibitem{he2017mask}
He, K., Gkioxari, G., Doll{\'a}r, P., Girshick, R.: Mask r-cnn. In: Proceedings
  of the IEEE international conference on computer vision. pp. 2961--2969
  (2017)

\bibitem{maskrcnn}
He, K., Gkioxari, G., Doll{\'{a}}r, P., Girshick, R.B.: Mask {R-CNN}. CoRR
  \textbf{abs/1703.06870} (2017), \url{http://arxiv.org/abs/1703.06870}

\bibitem{resnet}
He, K., Zhang, X., Ren, S., Sun, J.: Deep residual learning for image
  recognition. CoRR  \textbf{abs/1512.03385} (2015),
  \url{http://arxiv.org/abs/1512.03385}

\bibitem{bagoftricks}
He, T., Zhang, Z., Zhang, H., Zhang, Z., Xie, J., Li, M.: Bag of tricks for
  image classification with convolutional neural networks. In: The IEEE
  Conference on Computer Vision and Pattern Recognition (CVPR) (June 2019)

\bibitem{knowledge}
Hinton, G.E., Vinyals, O., Dean, J.: Distilling the knowledge in a neural
  network. ArXiv  \textbf{abs/1503.02531} (2015)

\bibitem{se-block}
{Hu}, J., {Shen}, L., {Sun}, G.: Squeeze-and-excitation networks. In: 2018
  IEEE/CVF Conference on Computer Vision and Pattern Recognition. pp.
  7132--7141 (June 2018). \doi{10.1109/CVPR.2018.00745}

\bibitem{stochastic_path}
Huang, G., Sun, Y., Liu, Z., Sedra, D., Weinberger, K.Q.: Deep networks with
  stochastic depth. CoRR  \textbf{abs/1603.09382} (2016),
  \url{http://arxiv.org/abs/1603.09382}

\bibitem{detection}
Huang, J., Rathod, V., Sun, C., Zhu, M., Korattikara, A., Fathi, A., Fischer,
  I., Wojna, Z., Song, Y., Guadarrama, S., Murphy, K.: Speed/accuracy
  trade-offs for modern convolutional object detectors. In: The IEEE Conference
  on Computer Vision and Pattern Recognition (CVPR) (July 2017)

\bibitem{jang2019learning}
Jang, Y., Lee, H., Hwang, S.J., Shin, J.: Learning what and where to transfer.
  In: International Conference on Machine Learning. pp. 3030--3039. PMLR (2019)

\bibitem{better_imagenet}
Kornblith, S., Shlens, J., Le, Q.V.: Do better imagenet models transfer better?
  In: The IEEE Conference on Computer Vision and Pattern Recognition (CVPR)
  (June 2019)

\bibitem{mimicking}
Li, Q., Jin, S., Yan, J.: Mimicking very efficient network for object
  detection. In: The IEEE Conference on Computer Vision and Pattern Recognition
  (CVPR) (July 2017)

\bibitem{crnet}
Liang, F., Lin, C., Guo, R., Sun, M., Wu, W., Yan, J., Ouyang, W.: Computation
  reallocation for object detection. In: International Conference on Learning
  Representations (2020), \url{https://openreview.net/forum?id=SkxLFaNKwB}

\bibitem{FPN}
{Lin}, T., {Dollár}, P., {Girshick}, R., {He}, K., {Hariharan}, B.,
  {Belongie}, S.: Feature pyramid networks for object detection. In: 2017 IEEE
  Conference on Computer Vision and Pattern Recognition (CVPR). pp. 936--944
  (July 2017). \doi{10.1109/CVPR.2017.106}

\bibitem{lin2014microsoft}
Lin, T.Y., Maire, M., Belongie, S., Hays, J., Perona, P., Ramanan, D.,
  Doll{\'a}r, P., Zitnick, C.L.: Microsoft coco: Common objects in context. In:
  European conference on computer vision. pp. 740--755. Springer (2014)

\bibitem{darts}
Liu, H., Simonyan, K., Yang, Y.: {DARTS}: Differentiable architecture search.
  In: International Conference on Learning Representations (2019),
  \url{https://openreview.net/forum?id=S1eYHoC5FX}

\bibitem{ensemble_resampling}
{Liu}, X., {Liu}, Z., {Wang}, G., {Cai}, Z., {Zhang}, H.: Ensemble transfer
  learning algorithm. IEEE Access  \textbf{6},  2389--2396 (2018).
  \doi{10.1109/ACCESS.2017.2782884}

\bibitem{mmd2}
Long, M., Wang, J.: Learning transferable features with deep adaptation
  networks. CoRR  \textbf{abs/1502.02791} (2015),
  \url{http://arxiv.org/abs/1502.02791}

\bibitem{mmd3}
Long, M., Wang, J.: Learning transferable features with deep adaptation
  networks. CoRR  \textbf{abs/1502.02791} (2015),
  \url{http://arxiv.org/abs/1502.02791}

\bibitem{gridrcnn}
Lu, X., Li, B., Yue, Y., Li, Q., Yan, J.: Grid r-cnn. In: The IEEE Conference
  on Computer Vision and Pattern Recognition (CVPR) (June 2019)

\bibitem{luo2016understanding}
Luo, W., Li, Y., Urtasun, R., Zemel, R.: Understanding the effective receptive
  field in deep convolutional neural networks. In: Advances in neural
  information processing systems. pp. 4898--4906 (2016)

\bibitem{instagram}
Mahajan, D., Girshick, R.B., Ramanathan, V., He, K., Paluri, M., Li, Y.,
  Bharambe, A., van~der Maaten, L.: Exploring the limits of weakly supervised
  pretraining. CoRR  \textbf{abs/1805.00932} (2018),
  \url{http://arxiv.org/abs/1805.00932}

\bibitem{evolution4}
Miikkulainen, R., Liang, J.Z., Meyerson, E., Rawal, A., Fink, D., Francon, O.,
  Raju, B., Shahrzad, H., Navruzyan, A., Duffy, N., Hodjat, B.: Evolving deep
  neural networks. CoRR  \textbf{abs/1703.00548} (2017),
  \url{http://arxiv.org/abs/1703.00548}

\bibitem{survey2}
{Pan}, S.J., {Yang}, Q.: A survey on transfer learning. IEEE Transactions on
  Knowledge and Data Engineering  \textbf{22}(10),  1345--1359 (Oct 2010).
  \doi{10.1109/TKDE.2009.191}

\bibitem{boosting2}
Pardoe, D., Stone, P.: Boosting for regression transfer. In: Proceedings of the
  27th International Conference on International Conference on Machine
  Learning. pp. 863--870. ICML'10, Omnipress, USA (2010),
  \url{http://dl.acm.org/citation.cfm?id=3104322.3104432}

\bibitem{pytorch}
Paszke, A., Gross, S., Chintala, S., Chanan, G., Yang, E., DeVito, Z., Lin, Z.,
  Desmaison, A., Antiga, L., Lerer, A.: Automatic differentiation in pytorch
  (2017)

\bibitem{weightshare}
Pham, H., Guan, M.Y., Zoph, B., Le, Q.V., Dean, J.: Efficient neural
  architecture search via parameter sharing. CoRR  \textbf{abs/1802.03268}
  (2018), \url{http://arxiv.org/abs/1802.03268}

\bibitem{evolution3}
Real, E., Aggarwal, A., Huang, Y., Le, Q.V.: Regularized evolution for image
  classifier architecture search (2018),
  \url{https://arxiv.org/pdf/1802.01548.pdf}

\bibitem{evolution2}
Real, E., Moore, S., Selle, A., Saxena, S., Suematsu, Y.L., Tan, J., Le, Q.V.,
  Kurakin, A.: Large-scale evolution of image classifiers. In: Proceedings of
  the 34th International Conference on Machine Learning - Volume 70. pp.
  2902--2911. ICML'17, JMLR.org (2017),
  \url{http://dl.acm.org/citation.cfm?id=3305890.3305981}

\bibitem{ren2015faster}
Ren, S., He, K., Girshick, R., Sun, J.: Faster r-cnn: Towards real-time object
  detection with region proposal networks. In: Advances in neural information
  processing systems. pp. 91--99 (2015)

\bibitem{faster-rcnn}
Ren, S., He, K., Girshick, R.B., Sun, J.: Faster {R-CNN:} towards real-time
  object detection with region proposal networks. CoRR  \textbf{abs/1506.01497}
  (2015), \url{http://arxiv.org/abs/1506.01497}

\bibitem{mobilenetv2}
Sandler, M., Howard, A.G., Zhu, M., Zhmoginov, A., Chen, L.: Inverted residuals
  and linear bottlenecks: Mobile networks for classification, detection and
  segmentation. CoRR  \textbf{abs/1801.04381} (2018),
  \url{http://arxiv.org/abs/1801.04381}

\bibitem{sun2018multi}
Sun, M., Yuan, Y., Zhou, F., Ding, E.: Multi-attention multi-class constraint
  for fine-grained image recognition. In: Proceedings of the European
  Conference on Computer Vision (ECCV). pp. 805--821 (2018)

\bibitem{survey}
Tan, C., Sun, F., Kong, T., Zhang, W., Yang, C., Liu, C.: A survey on deep
  transfer learning. CoRR  \textbf{abs/1808.01974} (2018),
  \url{http://arxiv.org/abs/1808.01974}

\bibitem{mnas}
Tan, M., Chen, B., Pang, R., Vasudevan, V., Sandler, M., Howard, A., Le, Q.V.:
  Mnasnet: Platform-aware neural architecture search for mobile. In: The IEEE
  Conference on Computer Vision and Pattern Recognition (CVPR) (June 2019)

\bibitem{efficentnet}
Tan, M., Le, Q.V.: Efficientnet: Rethinking model scaling for convolutional
  neural networks. CoRR  \textbf{abs/1905.11946} (2019),
  \url{http://arxiv.org/abs/1905.11946}

\bibitem{adv3}
Tzeng, E., Hoffman, J., Darrell, T., Saenko, K.: Simultaneous deep transfer
  across domains and tasks. In: The IEEE International Conference on Computer
  Vision (ICCV) (December 2015)

\bibitem{adv4}
Tzeng, E., Hoffman, J., Saenko, K., Darrell, T.: Adversarial discriminative
  domain adaptation. In: The IEEE Conference on Computer Vision and Pattern
  Recognition (CVPR) (July 2017)

\bibitem{mmd1}
Tzeng, E., Hoffman, J., Zhang, N., Saenko, K., Darrell, T.: Deep domain
  confusion: Maximizing for domain invariance. CoRR  \textbf{abs/1412.3474}
  (2014), \url{http://arxiv.org/abs/1412.3474}

\bibitem{WahCUB_200_2011}
Wah, C., Branson, S., Welinder, P., Perona, P., Belongie, S.: {The Caltech-UCSD
  Birds-200-2011 Dataset}. Tech. Rep. CNS-TR-2011-001, California Institute of
  Technology (2011)

\bibitem{boosting3}
Wang, C., Wu, Y., Liu, Z.: Hierarchical boosting for transfer learning with
  multi-source. In: Proceedings of the International Conference on Artificial
  Intelligence and Robotics and the International Conference on Automation,
  Control and Robotics Engineering. pp. 15:1--15:5. ICAIR-CACRE '16, ACM, New
  York, NY, USA (2016). \doi{10.1145/2952744.2952756},
  \url{http://doi.acm.org/10.1145/2952744.2952756}

\bibitem{non-local}
Wang, X., Girshick, R., Gupta, A., He, K.: Non-local neural networks. In: The
  IEEE Conference on Computer Vision and Pattern Recognition (CVPR) (June 2018)

\bibitem{WelinderEtal2010}
Welinder, P., Branson, S., Mita, T., Wah, C., Schroff, F., Belongie, S.,
  Perona, P.: {Caltech-UCSD Birds 200}. Tech. Rep. CNS-TR-2010-001, California
  Institute of Technology (2010)

\bibitem{fbnet}
Wu, B., Dai, X., Zhang, P., Wang, Y., Sun, F., Wu, Y., Tian, Y., Vajda, P.,
  Jia, Y., Keutzer, K.: Fbnet: Hardware-aware efficient convnet design via
  differentiable neural architecture search. In: The IEEE Conference on
  Computer Vision and Pattern Recognition (CVPR) (June 2019)

\bibitem{evolution}
Xie, L., Yuille, A.: Genetic cnn. In: The IEEE International Conference on
  Computer Vision (ICCV) (Oct 2017)

\bibitem{resnext}
Xie, S., Girshick, R., Dollar, P., Tu, Z., He, K.: Aggregated residual
  transformations for deep neural networks. In: The IEEE Conference on Computer
  Vision and Pattern Recognition (CVPR) (July 2017)

\bibitem{snas}
Xie, S., Zheng, H., Liu, C., Lin, L.: {SNAS}: stochastic neural architecture
  search. In: International Conference on Learning Representations (2019),
  \url{https://openreview.net/forum?id=rylqooRqK7}

\bibitem{metriclearningreweighting}
{Xu}, Y., {Pan}, S.J., {Xiong}, H., {Wu}, Q., {Luo}, R., {Min}, H., {Song}, H.:
  A unified framework for metric transfer learning. IEEE Transactions on
  Knowledge and Data Engineering  \textbf{29}(6),  1158--1171 (June 2017).
  \doi{10.1109/TKDE.2017.2669193}

\bibitem{yue2018compact}
Yue, K., Sun, M., Yuan, Y., Zhou, F., Ding, E., Xu, F.: Compact generalized
  non-local network. In: Advances in Neural Information Processing Systems. pp.
  6510--6519 (2018)

\bibitem{TCA}
Zhang, J., Li, W., Ogunbona, P.: Joint geometrical and statistical alignment
  for visual domain adaptation. In: The IEEE Conference on Computer Vision and
  Pattern Recognition (CVPR) (July 2017)

\bibitem{oldrl1}
Zoph, B., Le, Q.V.: Neural architecture search with reinforcement learning.
  CoRR  \textbf{abs/1611.01578} (2016), \url{http://arxiv.org/abs/1611.01578}

\bibitem{nasnet}
Zoph, B., Vasudevan, V., Shlens, J., Le, Q.V.: Learning transferable
  architectures for scalable image recognition. CoRR  \textbf{abs/1707.07012}
  (2017), \url{http://arxiv.org/abs/1707.07012}

\end{thebibliography}
\end{document}